\documentclass{article}
\usepackage{xspace}

\PassOptionsToPackage{numbers,compress}{natbib}
\usepackage[final]{neurips}

\usepackage{latexsym}
\usepackage{tcolorbox}

\usepackage[T1]{fontenc}
\usepackage[utf8]{inputenc}
\usepackage{microtype}

\usepackage{inconsolata}
\usepackage{wrapfig,lipsum,booktabs}
\usepackage{pifont}
\usepackage{makecell} %multiline cell
\usepackage{tabularx}
\usepackage[normalem]{ulem}
\useunder{\uline}{\ul}{}
\usepackage[export]{adjustbox}

\usepackage{hyperref}       % hyperlinks
\usepackage{url}            % simple URL typesetting
\usepackage{booktabs}       % professional-quality tables
\usepackage{amsfonts}       % blackboard math symbols
\usepackage{nicefrac}       % compact symbols for 1/2, etc.

\usepackage{graphicx}
\usepackage{subfigure}
\usepackage{float}
\usepackage{multirow}

\usepackage{listings}

\usepackage{amsmath}
\usepackage{amssymb}
\usepackage{mathtools}
\usepackage{amsthm}
\usepackage{balance}
\usepackage{flushend}

\usepackage{lipsum}
\usepackage{graphicx}
\usepackage{subcaption}
\usepackage{tikz}
\usetikzlibrary{shapes, arrows, positioning}

\usepackage{fancyhdr}

\hypersetup{
	colorlinks,
	citecolor=gray,
	linkcolor=red,
	urlcolor=blue}

\usepackage[capitalize,noabbrev]{cleveref}
\usepackage{tablefootnote}

\usepackage{color, colortbl,xcolor}
\usepackage{enumitem}
\usepackage{booktabs,arydshln}

\newcommand{\nbf}[1]{{\noindent \textbf{#1}}}

\makeatletter
\def\adl@drawiv#1#2#3{%
	\hskip.5\tabcolsep
	\xleaders#3{#2.5\@tempdimb #1{1}#2.5\@tempdimb}%
	#2\z@ plus1fil minus1fil\relax
	\hskip.5\tabcolsep}
\newcommand{\cdashlinelr}[1]{%
	\noalign{\vskip\aboverulesep
		\global\let\@dashdrawstore\adl@draw
		\global\let\adl@draw\adl@drawiv}
	\cdashline{#1}
	\noalign{\global\let\adl@draw\@dashdrawstore
		\vskip\belowrulesep}}
\makeatother

\setlist[itemize]{align=parleft,left=0pt..0.5em}
\setlist[enumerate]{align=parleft,left=0pt..0.5em}

\setlist[itemize]{align=parleft,left=0pt..0.8em}

\newcommand{\evomapheader}{%
  \noindent
  \begin{minipage}[c]{0.62\linewidth}
    \href{https://evomap.ai}{%
      \includegraphics[
        height=14pt,
        keepaspectratio
      ]{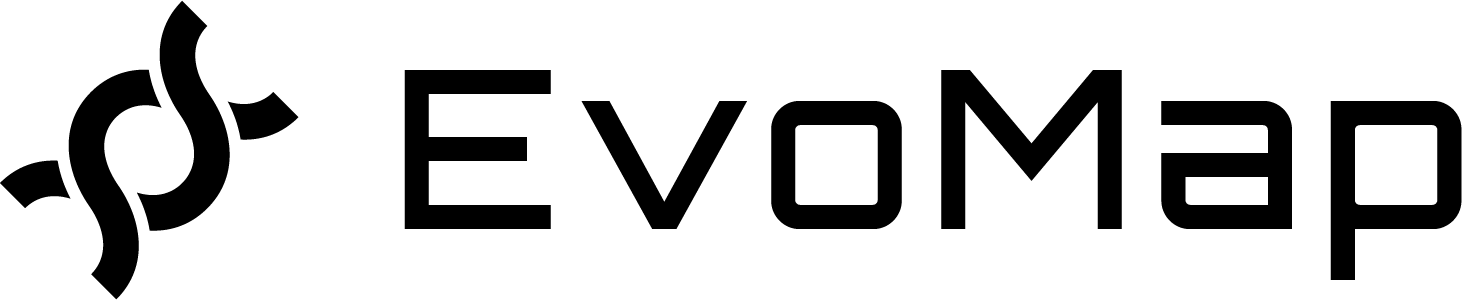}}
  \end{minipage}%
  \hfill
  \vspace{4pt}
}

\def\paperTitle{\includegraphics[height=14pt]{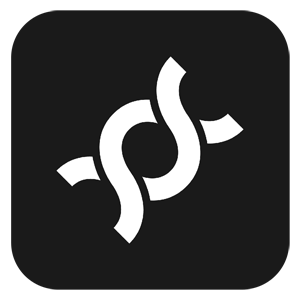} EvoX: Rethinking the Agent Harness as a Runtime Policy}

\title{\paperTitle}

\def\authorBlock{
    Haoyang Zhang\footnotemark[2] \qquad Qumeng Sun \qquad Xiao Zhang \qquad  Xiang Liu\\
    \textbf{Zhidong Fan \qquad Junjie Wang}
    \\\\\fontsize{10}{10}
    \selectfont{Infinite Evolution Lab, EvoMap}
    \\\\\fontsize{10}{10}
    \url{https://evomap.ai/evox}
}

\author{\authorBlock}

\begin{document}
\evomapheader
\vspace{-0.08in}
\maketitle

{
  \renewcommand{\thefootnote}%
  {\fnsymbol{footnote}}
  \footnotetext[2]{Corresponding Author. {\tt\small 17@evomap.ai}}
}

\begin{abstract}
Large language model (LLM) agents increasingly interact with tools and environments, yet plausible model decisions do not always translate into successful execution due to tool-call mismatches, interrupted workflows, and unsupported completion claims.
We present \texttt{EvoX}, a Rust-based, multi-model agent runtime guided by The Second Policy, which views the harness as an Operational Policy complementing the foundation model's Semantic Policy.
\texttt{EvoX} combines model-facing guidance with deterministic runtime control through three complementary mechanisms: Action Realization, Execution Continuity, and Outcome Grounding.
To evaluate harnesses without suppressing their native behaviors, we introduce \texttt{Contrast}, a controlled and auditable framework integrating explicit fairness constraints, provider-bound request verification, and evidence-based adjudication.
Across five benchmarks, six foundation models, and six agent harnesses, \texttt{EvoX} achieves the highest numerical success rate of 56.2\% over 691 common task--model pairs, exceeding the average of the other five harnesses by approximately 3.0 percentage points.
Its larger advantages on GLM-5.3 (+10.7 points) and Kimi K3 (+7.9 points) highlight the model-dependent effectiveness of harness design and the potential for cross-provider adaptability.
Achieved without memory, evolution, or extensions, these results establish \texttt{EvoX} as a competitive multi-model agent runtime and underscore the importance of harness-level operational control in realizing foundation-model capabilities.
\end{abstract}

\section{Introduction}
\label{sec:intro}

Recent advances in large language models (LLMs) have enabled AI agents to move beyond code generation toward autonomous tool use and multi-step execution in real-world environments~\citep{DBLP:journals/fcsc/WangMFZYZCTCLZWW24,LiuEtAl2026Agent4SE,DBLP:conf/coling/Li25,DBLP:conf/iclr/YaoZYDSN023}.
However, plausible model decisions do not guarantee successful execution, as tool-call mismatches, recoverable failures, and premature termination can prevent agents from completing their tasks.

Agent harnesses mediate this transition by connecting model decisions with tools, execution states, and environmental feedback.
Existing systems adopt different designs: SWE-agent emphasizes task-oriented agent--computer interfaces~\citep{DBLP:conf/nips/YangJWLYNP24}, OpenHands provides a general-purpose development environment~\citep{DBLP:conf/iclr/0001LSXTZPSLSTL25}, Claude Code and Codex support tool-driven coding workflows~\citep{Anthropic2026ClaudeCode,OpenAI2026CodexCLI}, while Pi adopts a minimal, extensible agent loop~\citep{EarendilWorks2026Pi}.
Despite these differences, harnesses influence not only tool access but also how models process feedback, recover from failures, and manage execution.
Recent studies further demonstrate that harness design can substantially affect task success and efficiency, even when the underlying model remains unchanged~\citep{DBLP:journals/corr/abs-2605-27922,VatsGolev2026ScaffoldEffect,Lewis2026DifferentHarness}.
This raises a fundamental question: \textit{What role should the harness play in turning model decisions into reliable execution?}

To make this role explicit, we introduce \texttt{EvoX}, a Rust-based, multi-model agent system guided by a design principle termed \textbf{The Second Policy}.
This principle distinguishes two complementary policies: a \textit{Semantic Policy}, embodied by the foundation model, that interprets tasks, plans actions, and reasons over feedback; and an \textit{Operational Policy}, embodied by the harness, that governs action execution, runtime states, and recovery.
Rather than introducing a separately trained model, the operational policy combines model-facing guidance with deterministic runtime controls to mediate execution without modifying foundation model parameters.

\texttt{EvoX} realizes this design through three complementary mechanisms.
\textit{Action Realization} detects malformed or text-leaked tool calls, validates tool schemas, and normalizes compatible arguments to mitigate mismatches between model outputs and executable tools.
\textit{Execution Continuity} combines failure-recovery guidance, repeated-error detection, runtime circuit breakers, and context management to support recovery while limiting unproductive execution loops.
\textit{Outcome Grounding} guides models to align reported outcomes with observed actions and tool feedback.
Beyond its core harness, \texttt{EvoX} supports diverse models, tools, and multi-agent workflows, alongside optional long-term memory and Gene Evolution Protocol (GEP)-based experience evolution~\citep{DBLP:journals/corr/abs-2604-15097}.

Evaluating these harness designs introduces a related challenge: ensuring comparable execution conditions without suppressing the native behaviors that distinguish individual harnesses.
We therefore develop \texttt{Contrast}, a controlled and auditable framework for cross-harness evaluation.
\texttt{Contrast} distinguishes locked, shared, and harness-native conditions, integrating task qualification, model-request verification, execution evidence, and failure-aware adjudication to support systematic and traceable comparisons.

Using \texttt{Contrast}, we evaluate \texttt{EvoX} against five agent harnesses across five benchmarks and six foundation models under controlled conditions.
Across 691 common task--model pairs, \texttt{EvoX} achieves the highest numerical success rate of 56.2\%, followed by Pi (55.9\%), Claude Code (55.3\%), and Hermes (54.4\%).
It exceeds the average of the other five harnesses by approximately 3.0 percentage points, with larger advantages of 10.7 and 7.9 points on GLM-5.3 and Kimi K3, respectively.
The results reveal model-dependent harness effectiveness: Claude Code and Codex lead on their affiliated Opus 5 and GPT-6 Astra models, whereas \texttt{EvoX} leads on GLM-5.3 and Kimi K3, demonstrating competitive cross-provider adaptability.
Achieved without memory, evolution, or extensions, these findings position \texttt{EvoX} as a competitive multi-model agent runtime and highlight the importance of harness-level \emph{Operational Policy} in realizing foundation-model capabilities.

\section{Related Work}
\label{sec:related}

\subsection{Agent Harnesses and Runtime Control}

Agent harnesses shape how model-generated decisions are translated into executable actions through their choices of interfaces, tools, and runtime control.
SWE-agent~\citep{DBLP:conf/nips/YangJWLYNP24} introduces specialized agent--computer interfaces for repository navigation, code editing, and testing, while CodeAct~\citep{WangEtAl2024CodeAct} uses executable code as a compositional action space.
OpenHands~\citep{DBLP:conf/iclr/0001LSXTZPSLSTL25} provides sandboxed execution and multi-agent coordination, whereas Claude Code and Codex integrate tool orchestration and execution controls into development workflows~\citep{Anthropic2026ClaudeCode,OpenAI2026CodexCLI}.
Other harnesses explore different runtime architectures: Pi emphasizes a minimal, extensible agent loop~\citep{EarendilWorks2026Pi}, Hermes Agent incorporates persistent memory and reusable skills~\citep{NousResearch2026HermesAgent}, and Prime Agent employs a persistent execution environment with recursive subagents and continual harness state~\citep{KartenEtAl2026PrimeAgent}.
Complementary approaches improve reliability through feedback and structured workflows: Reflexion~\citep{Shinn2023Reflexion} uses verbal feedback for subsequent attempts, CRITIC~\citep{Gou2024CRITIC} performs tool-assisted self-correction, and Agentless~\citep{Xia2025Agentless} structures software repair into localization, patch generation, and validation.
These approaches demonstrate the importance of execution design beyond model reasoning, while addressing different aspects of agent behavior.
Building on these insights, \texttt{EvoX} introduces ``The Second Policy'' to organize action realization, execution continuity, and outcome grounding as complementary responsibilities of harness-level operational control.

\subsection{Controlled Evaluation of Agent Harnesses}

Reliable agent evaluation requires both executable task verification and comparable execution conditions.
SWE-bench~\citep{Jimenez2024SWEBench} evaluates repository-level issue resolution, SWE-bench Live~\citep{Zhang2025SWEBenchLive} addresses benchmark freshness and contamination, and AgentBoard~\citep{Ma2024AgentBoard} introduces fine-grained progress measurements beyond final success.
More directly, Harness-Bench~\citep{DBLP:journals/corr/abs-2605-27922} and The Scaffold Effect~\citep{VatsGolev2026ScaffoldEffect} investigate harness-dependent performance, while Same Model, Different Harness~\citep{Lewis2026DifferentHarness} examines context management and stalled-execution handling under fixed model weights.
However, identical tasks and nominal model configurations do not necessarily ensure comparable execution when request parameters, runtime conditions, and failure attribution differ.
To address these challenges, \texttt{Contrast} distinguishes locked, shared, and harness-native conditions, integrating task qualification, model-request verification, and evidence-based adjudication to support controlled comparisons.

\begin{figure*}[!t]
\centering
\includegraphics[width=\textwidth]{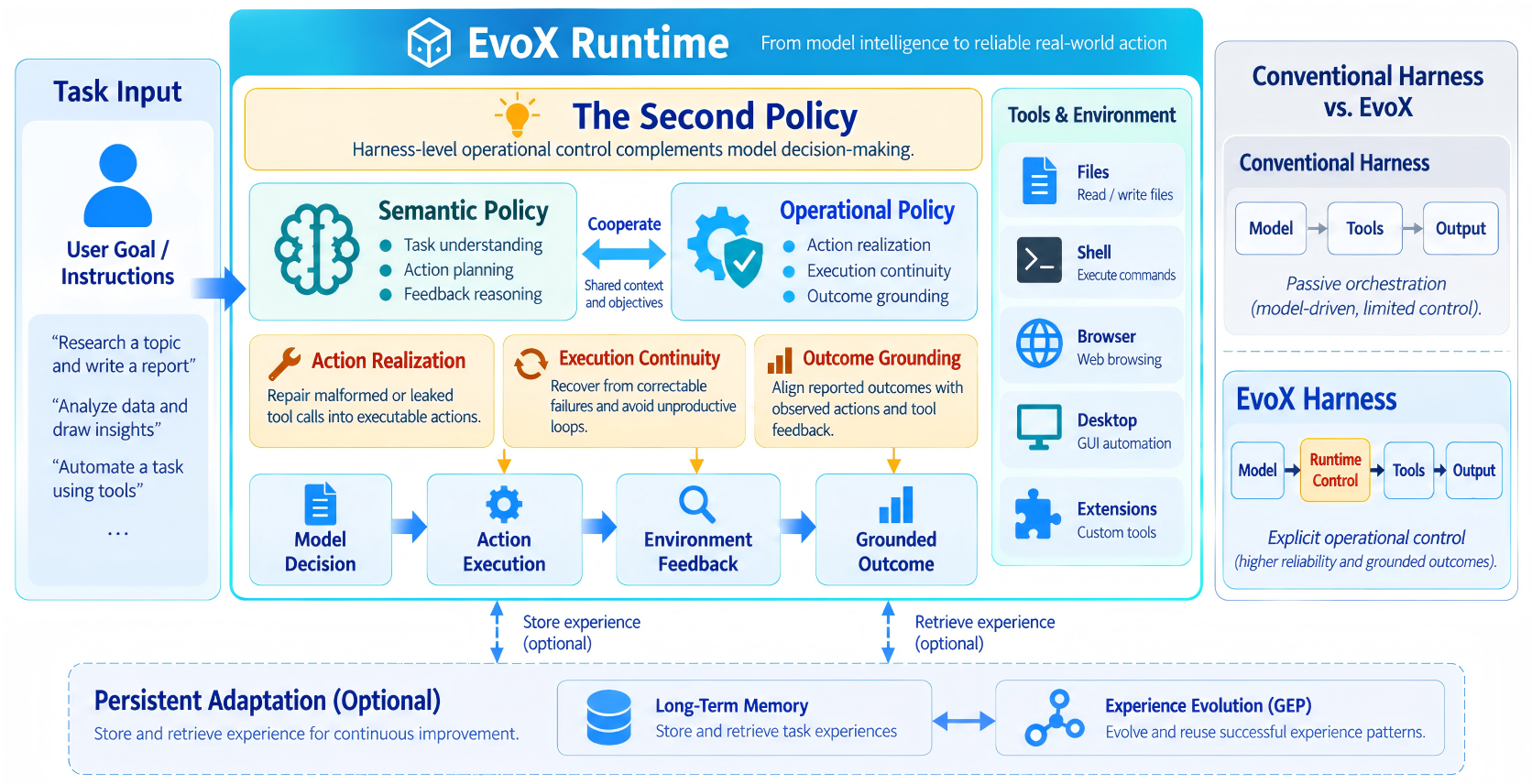}
\caption{
\textbf{Architecture and runtime design of \texttt{EvoX}.}
The system integrates a model-driven \emph{Semantic Policy} with a harness-level \emph{Operational Policy} to coordinate decisions and environment interactions.
The operational policy comprises three complementary mechanisms:
\emph{Action Realization} mediates executable tool calls,
\emph{Execution Continuity} coordinates recovery and runtime control,
and \emph{Outcome Grounding} aligns reported outcomes with observed execution.
The runtime supports diverse tools and optional persistent adaptation through long-term memory and GEP-based experience evolution.
The right panel illustrates the conceptual distinction between a minimal execution loop and explicit harness-level operational control.
}
\label{fig:evox-architecture}
\end{figure*}

\section{\texttt{EvoX}: Architecture and Runtime Design}
\label{sec:evox}

\subsection{System Overview and Execution Flow}
\label{sec:evox-overview}

\nbf{System Architecture.}
\texttt{EvoX} is a Rust-based, multi-model agent runtime that connects foundation models with tools and interactive environments through a unified execution framework.
As illustrated in~\cref{fig:evox-architecture}, the system integrates user-facing interfaces, model invocation, context management, tool execution, and runtime state management.
Its architecture distinguishes model-driven semantic decisions from harness-level operational control, establishing the foundation for \textbf{The Second Policy}.
The runtime accommodates heterogeneous models, tools, and interaction interfaces, with additional capabilities for multi-agent coordination and persistent adaptation.

\nbf{Execution Flow.}
Given a user task $x$, \texttt{EvoX} maintains an interaction history $h_t$ and a runtime state $s_t$ at each interaction step $t$.
The history $h_t$ records model outputs and environmental feedback, while $s_t$ tracks execution conditions, including task status and failure information.
The system assembles the model context from task instructions, interaction history, available tool interfaces $\mathcal{T}_t$, and relevant runtime information.
The model then proposes an action, which the harness interprets and dispatches when executable under the applicable controls.
Tool results update the interaction history and runtime state, forming an iterative decision--execution--feedback loop.
Crucially, the realized execution depends not only on the model's proposal but also on how the harness processes it.

\subsection{The Second Policy}
\label{sec:evox-policy}

The distinction between proposed actions and realized execution reveals a fundamental role of the agent harness.
A semantically appropriate action may fail to satisfy an executable interface, recoverable errors may interrupt task progress, and final responses may overstate what execution has established.
Rather than treating these issues as isolated execution exceptions, \texttt{EvoX} organizes their handling through a unified design principle termed \textbf{The Second Policy}.
This perspective recognizes the harness as an explicit source of operational control that complements, rather than replaces, model-driven decision-making.

\nbf{Complementary Decision Policies.}
The \emph{Semantic Policy}, represented by the foundation model, interprets task objectives, proposes actions, and revises decisions according to interaction feedback.
The \emph{Operational Policy}, implemented by the harness, governs how these proposals are interpreted, executed, and incorporated into subsequent interactions.
While semantic decisions determine what the agent attempts, operational control manages how those decisions are realized under runtime constraints.
Importantly, the harness influences both the context in which actions are proposed and the conditions under which they are executed.

We formalize this interaction by representing the semantic policy as $\pi_\theta$ and the operational policy as $\mathcal{H}=(\Gamma,\Delta)$, where $\Gamma$ denotes model-facing guidance and $\Delta$ denotes deterministic runtime control.
At each interaction step, the two policies cooperate as
\begin{equation}
\label{eq:evox-policy}
\begin{aligned}
g_t &= \Gamma(h_t,s_t),\\
c_t &= \operatorname{Ctx}(x,h_t,g_t,\mathcal{T}_t),\\
u_t &\sim \pi_\theta(\cdot\mid c_t),\\
(\hat{a}_t,d_t) &= \Delta(u_t,s_t,\mathcal{T}_t).
\end{aligned}
\end{equation}
Here, $\Gamma$ supplies runtime guidance $g_t$, which is incorporated into the model context $c_t$ before generating an output $u_t$.
The runtime controller $\Delta$ subsequently interprets this output, producing an executable action $\hat{a}_t$ when applicable and a disposition $d_t$ indicating how execution should proceed.
Depending on the proposal and runtime conditions, the disposition may permit execution, return a diagnostic, or terminate the interaction.
The resulting tool feedback or runtime diagnostic $o_t$ updates $h_{t+1}$ and $s_{t+1}$, informing subsequent model decisions and runtime control.
This formulation describes complementary responsibilities within the agent system rather than introducing an additional trained policy network.

\nbf{Hybrid Operational Control.}
The two components of $\mathcal{H}$ address different aspects of execution reliability.
Model-facing guidance $\Gamma$ incorporates behavioral expectations and observed feedback into subsequent model interactions, encouraging error recovery and evidence-grounded reporting while preserving flexibility in semantic decisions.
In contrast, runtime control $\Delta$ applies explicit rules to action representations and execution states, including tool validation, bounded interface correction, and repeated-error detection.
Such interventions may reject incompatible operations or terminate unproductive execution paths without requiring additional model decisions.
Accordingly, guidance shapes how the model responds to execution feedback, while deterministic control governs specific runtime conditions where applicable.
Together, they establish operational control across the interaction loop rather than at a single execution stage.

\subsection{Harness-Level Operational Mechanisms}
\label{sec:evox-mechanisms}

The operational policy is realized through three complementary mechanisms that address action execution, interaction continuity, and outcome reporting.
Each mechanism operates at a distinct point in the execution loop while sharing the control formulation in~\cref{eq:evox-policy}.

\nbf{Action Realization.}
A model may select an appropriate tool yet produce an invocation that the execution interface cannot directly process.
Given a model output $u_t$, Action Realization uses runtime control $\Delta$ to examine its compatibility with registered tool interfaces through tool-call detection, schema validation, and argument normalization.
Where supported, this includes identifying malformed structured calls or recognizable tool intents expressed in nonstandard textual forms.
When a compatible representation can be recovered, the runtime applies bounded normalization to obtain an executable action $\hat{a}_t$.
Otherwise, the invocation is rejected or returned for model-side correction rather than being arbitrarily reinterpreted.
This mechanism separates semantic action intent from executable representation, reducing avoidable interface-level failures while preserving the model's intended operation.

\nbf{Execution Continuity.}
Recoverable execution failures, such as invalid paths, missing dependencies, or unsuccessful commands, need not imply that the underlying task is infeasible.
After receiving tool feedback $o_t$, the harness updates the interaction history $h_{t+1}$ and runtime state $s_{t+1}$ to reflect the observed execution.
Model-facing guidance $\Gamma$ then encourages the semantic policy to diagnose failures and revise subsequent actions rather than prematurely terminating the task.
Meanwhile, runtime control $\Delta$ monitors recurring tool-error patterns and, when enabled, applies circuit breakers to terminate repeated, unproductive attempts.
Context management and automatic compaction further support extended interactions by maintaining a manageable history as context limits are approached.
Execution Continuity therefore balances flexible, model-guided recovery with state-dependent stopping decisions, allowing execution to proceed when correction remains possible while limiting ineffective repetition.

\nbf{Outcome Grounding.}
A model-generated completion claim does not necessarily establish that the corresponding operations were successfully executed.
When producing a final response, the semantic policy is guided by $\Gamma$ to distinguish intended actions from completed operations and to ground its claims in the observations recorded in $h_t$.
\texttt{EvoX} incorporates grounded-action guidance that encourages the model to report task outcomes according to observed tool results rather than unsupported assumptions of success.
Unlike deterministic action validation or circuit breaking, this mechanism primarily operates through model-facing instructions rather than independent correctness verification.
Outcome Grounding therefore targets consistency between observed execution and model-generated reports, without treating those reports as proof of task completion.

Together, these mechanisms operationalize \textbf{The Second Policy} across action interpretation, execution continuity, and outcome reporting.
They preserve the foundation model's responsibility for semantic reasoning while making selected aspects of execution explicitly manageable by the harness.

\subsection{Extensibility and Persistent Adaptation}
\label{sec:evox-adaptation}

Beyond its core operational mechanisms, \texttt{EvoX} provides extensible execution capabilities and persistent experience reuse across tasks.

\nbf{Extensible Agent Infrastructure.}
\texttt{EvoX} supports multiple foundation model providers and diverse execution tools, including file and shell operations, web retrieval, browser interaction, and desktop control.
Its modular interfaces enable external integrations through the Model Context Protocol (MCP), reusable skills, and native extensions.
For complex tasks, the runtime supports subagent delegation, coordinated workflows, and optional swarm-based multi-agent collaboration.
These capabilities extend the same runtime architecture across heterogeneous environments and collaborative execution settings without modifying foundation model parameters.

\nbf{Persistent Adaptation.}
Beyond individual tasks, \texttt{EvoX} incorporates long-term memory and Gene Evolution Protocol (GEP)-based experience evolution~\citep{DBLP:journals/corr/abs-2604-15097}.
Long-term memory extracts reusable information from interaction histories and retrieves task-relevant memories to enrich subsequent model contexts.
GEP organizes execution experiences into \emph{Capsules}, from which recurring successful patterns can be consolidated into reusable \emph{Genes}.
Selected Genes provide task-specific guidance and may restrict available tools or influence model routing, allowing accumulated experience to inform future execution.
These capabilities complement within-task operational control with persistent, experience-informed adaptation.

\begin{figure*}[!t]
\centering
\includegraphics[width=\textwidth]{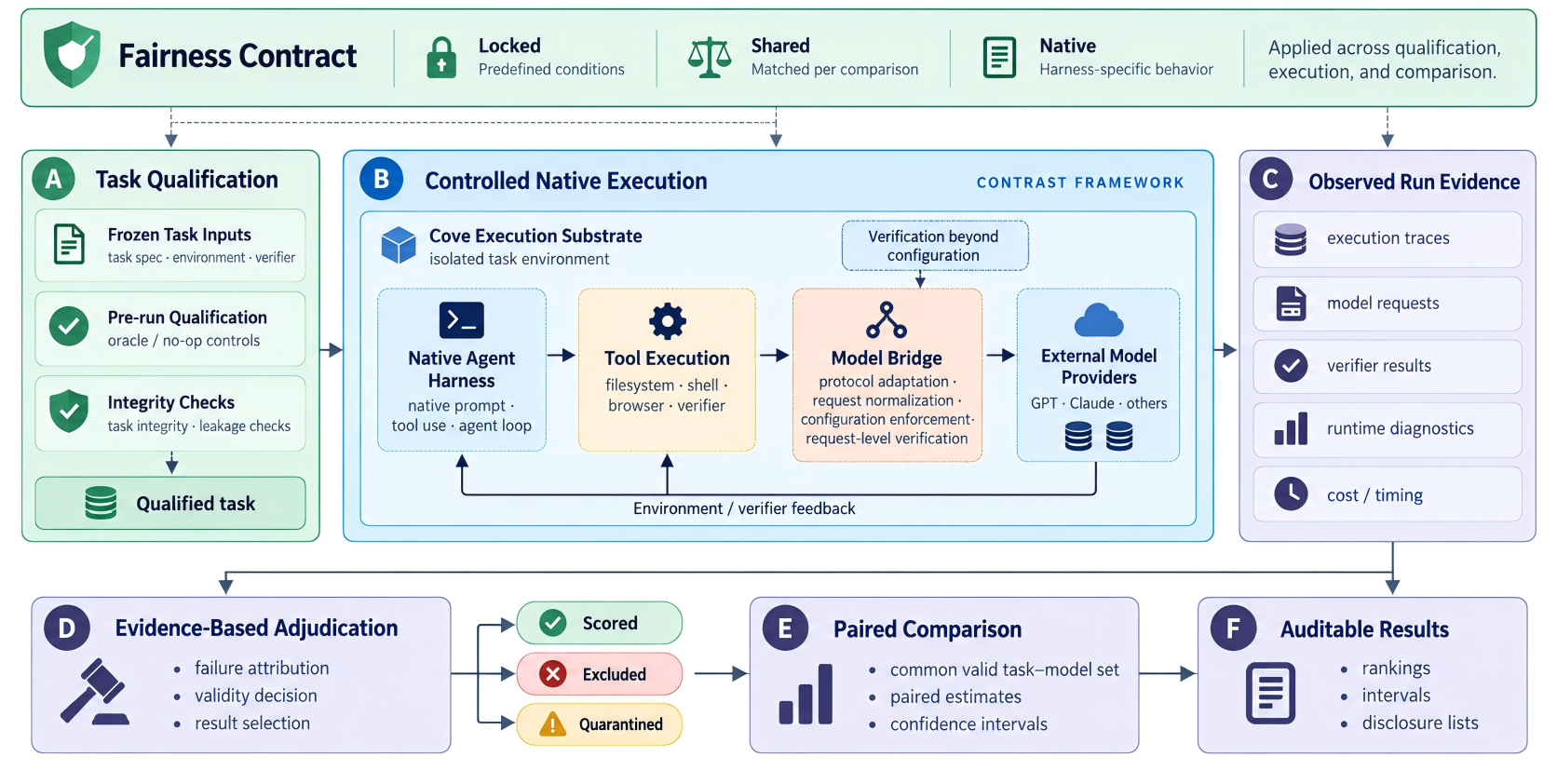}
\caption{
\textbf{Overview of \texttt{Contrast} for controlled and auditable harness evaluation.}
A fairness contract distinguishes locked, shared, and harness-native conditions, preserving native execution behaviors while controlling external evaluation factors.
Versioned tasks undergo qualification and integrity checks before execution through native harnesses in controlled environments.
A model bridge mediates model requests, enforces applicable configurations, and audits provider-bound request parameters.
Execution traces, model-call records, verifier results, and runtime diagnostics provide evidence for failure attribution, validity decisions, and auditable result selection.
Only valid, scored task--model combinations enter paired statistical comparisons, while excluded or quarantined records remain available for audit and disclosure.
}
\label{fig:contrast-framework}
\end{figure*}

\section{\texttt{Contrast}: Controlled and Auditable Harness Evaluation}
\label{sec:contrast}

\subsection{Evaluation Principles and Fairness Contract}
\label{sec:contrast-fairness}

Evaluating agent harnesses presents a fundamental challenge: external execution conditions must remain comparable without eliminating the native behaviors that distinguish individual harnesses.
Standardizing internal prompts, tools, and control loops would obscure the mechanisms under study, whereas leaving external conditions uncontrolled could confound performance comparisons.
Inspired by Harbor's unified framework for evaluating heterogeneous agents~\citep{Harbor_Framework}, we introduce \texttt{Contrast}, a controlled and auditable evaluation framework built around an explicit \emph{fairness contract}.
As illustrated in~\cref{fig:contrast-framework}, this contract governs task qualification, harness execution, model-call verification, and subsequent result comparison.

\nbf{Evaluation Objective.}
Let $\tau_i$, $m_k$, and $a_j$ denote an evaluation task, a foundation model, and an agent harness, respectively.
We represent the resulting task--model--harness execution as
\begin{equation}
\label{eq:contrast-evaluation}
r_{ijk}
=
\operatorname{Run}
\left(
\tau_i,m_k,a_j;
\boldsymbol{\phi}_{ik},\boldsymbol{\nu}_{ijk}
\right),
\end{equation}
where $\boldsymbol{\phi}_{ik}$ specifies externally controlled evaluation conditions and $\boldsymbol{\nu}_{ijk}$ represents harness-native configurations and behaviors.
The outcome $r_{ijk}$ includes the execution result and associated runtime evidence, which are subsequently examined for validity and comparability.
The objective is to compare harnesses over matched task--model combinations while preserving their intrinsic execution strategies, rather than enforcing identical internal trajectories.

\nbf{Explicit Fairness Contract.}
\texttt{Contrast} classifies evaluation conditions into three complementary categories: \emph{locked}, \emph{shared}, and \emph{native}.
\emph{Locked conditions} follow predefined rules, including model-compatible reasoning configurations, output limits, and network access policies.
\emph{Shared conditions} maintain consistent values within each comparison, such as model identity, provider routing, and execution budgets.
In contrast, \emph{native conditions} preserve harness-specific properties, including system prompts, tool interfaces, context management, and internal recovery strategies.
These native differences are recorded rather than homogenized, allowing harness design to remain the primary variable of interest.
The fairness contract therefore establishes comparability through explicit control of external factors without suppressing the operational characteristics being evaluated.

\nbf{Cross-Stage Verification.}
Importantly, comparable configurations do not necessarily imply equivalent execution conditions.
\texttt{Contrast} consequently verifies applicable constraints at multiple stages, from pre-execution configuration checks to inspection of model requests forwarded to providers.
For supported model-call routes, controlled parameters are enforced and audited against the fairness contract, while harness-native behaviors remain observable through execution records.
This approach distinguishes declared configurations from realized model-call conditions and supports subsequent evidence-based validity assessment.
The framework connects controlled experimental design with auditable execution, providing a consistent basis for comparing heterogeneous agent harnesses.

\subsection{Task Qualification and Integrity}
\label{sec:contrast-tasks}

Reliable harness comparison requires consistent task instances and verifiable evaluation outcomes.
As shown in~\cref{fig:contrast-framework}(A), \texttt{Contrast} qualifies benchmark tasks before execution to reduce inconsistencies in task artifacts, verifier behavior, and answer exposure.

\nbf{Task Standardization and Identity.}
\texttt{Contrast} materializes heterogeneous benchmarks into a common task specification comprising instructions, execution environments, and task-specific verifiers.
Each task $\tau_i$ is associated with versioned artifacts and content-based identities that link qualification evidence to subsequent execution.
Task contents, relevant file permissions, and environment identities are checked against prepared configurations to detect unintended changes.
Benchmark-specific capability requirements are also considered to identify incompatible execution interfaces without altering harness-native behaviors.

\nbf{Qualification and Integrity Checks.}
\texttt{Contrast} employs complementary positive and negative controls to assess basic verifier reliability.
An \emph{Oracle} applies the reference solution to confirm that successful execution is recognized, while a \emph{No-op} checks that an unchanged task state is not incorrectly rewarded.
Qualification evidence is bound to task identities, preventing modified instances from inheriting previous validation.
Additional integrity checks examine task environments, cached artifacts, and accessible resources for potential exposure of reference solutions or hidden tests.
Tasks with unresolved concerns are withheld pending review, while post-run evidence is further examined for contamination during adjudication (\cref{sec:contrast-adjudication}).

\subsection{Controlled Native Execution and Model-Call Verification}
\label{sec:contrast-execution}

Establishing comparable evaluation conditions requires more than aligning declared configurations, as heterogeneous harnesses differ in execution interfaces, model protocols, and runtime behavior.
As illustrated in~\cref{fig:contrast-framework}(B), \texttt{Contrast} combines native harness execution with controlled environments and provider-bound model-call verification to operationalize the fairness contract.

\nbf{Harness-Native Execution.}
For each task--model--harness combination $(\tau_i,m_k,a_j)$, \texttt{Contrast} launches the corresponding agent through a standardized adapter while preserving its native system prompts, tool interfaces, execution loops, and recovery strategies.
The Cove execution substrate provides task-specific containers, controlled resources, and environment configurations, ensuring that participating harnesses operate against consistent evaluation targets.
Harness versions, task identities, and runtime configurations are recorded to establish execution provenance.
This design preserves the harness-specific behaviors represented by $\boldsymbol{\nu}_{ijk}$ while enforcing applicable external conditions specified by $\boldsymbol{\phi}_{ik}$.

\nbf{Provider-Bound Request Verification.}
Identical model identifiers and nominal inference settings do not necessarily produce equivalent API requests across harnesses.
To address this discrepancy, \texttt{Contrast} routes supported model calls through a shared \emph{Model Bridge} that adapts heterogeneous API protocols and enforces applicable constraints on model identity, reasoning configurations, and output limits.
Beyond pre-execution configuration checks, the framework inspects provider-bound requests to verify that controlled parameters are reflected in the actual transmitted payloads.
This separates declared settings from realized model-call conditions, reducing configuration-induced discrepancies without modifying native agent decision logic.
Request-level verification establishes compliance with observable protocol constraints but cannot guarantee identical provider-side execution semantics.

\nbf{Campaign Orchestration and Provenance.}
\texttt{Contrast} organizes task--model--harness combinations into controlled evaluation campaigns with explicit execution budgets, versioned configurations, and traceable trial identities.
Each run produces execution records linking agent actions, model requests, verifier outputs, runtime diagnostics, and resource consumption.
These records support subsequent validity checks and failure attribution rather than treating every unsuccessful run as an agent failure.
The resulting evidence is collected for the adjudication process described in~\cref{sec:contrast-adjudication}.

\subsection{Evidence-Based Adjudication and Paired Comparison}
\label{sec:contrast-adjudication}

Controlled execution alone does not guarantee reliable evaluation, as unsuccessful runs may arise from agent failures, infrastructure errors, or invalid assessment conditions.
To distinguish these cases, \texttt{Contrast} evaluates execution outcomes against recorded evidence before incorporating them into performance comparisons.
As illustrated in~\cref{fig:contrast-framework}(C--F), this process connects execution records, failure attribution, result selection, and paired statistical analysis.

\nbf{Execution Evidence.}
For each trial $r_{ijk}$, \texttt{Contrast} collects execution evidence $e_{ijk}$ comprising agent traces, provider-bound model requests, verifier outcomes, runtime diagnostics, and resource consumption.
These records are associated with task, model, harness, and execution identities, allowing reported outcomes to be traced to their underlying observations.
Evidence integrity and measurement completeness are examined to identify missing, inconsistent, or potentially contaminated records.
This supports the distinction between task-level outcomes and evaluation artifacts.

\nbf{Failure Attribution and Adjudication.}
As shown in~\cref{fig:contrast-framework}(D), \texttt{Contrast} applies evidence-based rules to determine whether an execution constitutes a valid task outcome.
Given an adjudication protocol $\rho$, we express this process as
\begin{equation}
\label{eq:contrast-adjudication}
(y_{ijk},q_{ijk})
=
\operatorname{Adjudicate}(r_{ijk},e_{ijk};\rho),
\end{equation}
where $y_{ijk}$ denotes the task score and $q_{ijk}$ indicates whether the record is scored, excluded, or quarantined.
Importantly, scored records include both successful executions and attributable agent failures, whereas verified evaluation-side faults may invalidate a trial.
Records with unresolved evidence conflicts can be isolated for further review rather than automatically treated as successful or excluded.
This distinction prevents infrastructure failures from being indiscriminately attributed to agents while retaining legitimate task failures in the evaluation.

\nbf{Auditable Result Selection.}
When multiple attempts exist for a task--model--harness combination, \texttt{Contrast} applies versioned selection and invalidation rules to determine its representative outcome.
The selected record retains its execution provenance, while superseded or excluded attempts remain traceable through the corresponding evidence.
This process separates result selection from observed task performance and supports consistent application of the evaluation protocol.
Subsequent comparisons use the selected scores and their associated validity decisions.

\nbf{Paired Comparison and Reporting.}
Rather than aggregating different sets of valid executions for each harness, \texttt{Contrast} constructs a common task--model set in which all compared harnesses have scored outcomes, as illustrated in~\cref{fig:contrast-framework}(E).
Formally, for the set of compared harnesses $\mathcal{A}$,
\begin{equation}
\label{eq:contrast-common-set}
\mathcal{I}_{\mathrm{common}}
=
\left\{
(i,k)\mid
q_{ijk}=\mathrm{scored},
\ \forall a_j\in\mathcal{A}
\right\}.
\end{equation}
Harness performance and pairwise differences are evaluated over this matched set, with uncertainty estimated through paired, task-level resampling.
The resulting reports include performance estimates, confidence intervals, and disclosure records for excluded or unresolved executions (\cref{fig:contrast-framework}(F)).
Together, these procedures establish an auditable measurement process in which performance comparisons are grounded in validated execution evidence rather than raw trial outcomes alone.

\section{Experiments and Analysis}
\label{sec:experiments}

\subsection{Evaluation Setup}
\label{sec:exp-setup}

We evaluate \texttt{EvoX} using \texttt{Contrast} to examine how different agent harnesses affect task outcomes under matched foundation-model and evaluation conditions.

\nbf{Evaluation Scope.}
We compare six agent harnesses: \texttt{EvoX}, Claude Code~\citep{Anthropic2026ClaudeCode}, Codex~\citep{OpenAI2026CodexCLI}, Hermes~\citep{NousResearch2026HermesAgent}, Pi~\citep{EarendilWorks2026Pi}, and Prime Agent~\citep{KartenEtAl2026PrimeAgent}.
The evaluation covers six foundation models: Opus 5~\citep{Anthropic2026ClaudeOpus5}, Sonnet 5~\citep{Anthropic2026ClaudeSonnet5}, GPT-6 Astra~\citep{OpenAI2026GPT6Astra}, GLM-5.3~\citep{ZAI2026GLM53}, Kimi K3~\citep{KimiTeam2026KimiK3}, and DeepSeek V4 Flash~\citep{DBLP:journals/corr/abs-2606-19348}.
We consider five benchmarks covering multilingual programming, web-assisted reasoning, algorithmic problem solving, repository-level software repair, and terminal-based engineering.
Given computational constraints, we aim to select approximately the first 28 eligible tasks from each benchmark's predefined evaluation panel, subject to task qualification and availability.
The resulting evaluation includes 28 tasks from Aider Polyglot~\citep{Aider2024Polyglot}, 21 from GAIA~\citep{DBLP:conf/iclr/MialonF0LS24}, 28 from Harbor Index~\citep{harbor_index}, 27 from SWE-Live~\citep{Zhang2025SWEBenchLive}, and 28 from Terminal-Bench 2.1~\citep{TerminalBenchTeam2026TB21}.
Together, these 132 tasks yield 4,752 planned task--model--harness combinations.

\nbf{Controlled Harness Configurations.}
Following the fairness contract in~\cref{sec:contrast-fairness}, we align foundation-model identities, model-specific reasoning configurations, requested output limits, execution budgets, and applicable network restrictions across harnesses.
Model requests are routed through a shared gateway, with controlled parameters checked against provider-bound requests.
Meanwhile, each harness retains its native system prompts, tool interfaces, context management, execution loops, and recovery strategies.
\texttt{EvoX} is evaluated in its non-interactive execution mode using version \texttt{v1.1.0-pre.93}, with memory, evolution, and extensions disabled.
This configuration focuses the comparison on baseline harness execution rather than optional persistent adaptation.

\nbf{Metrics.}
We measure task success using binary outcomes determined by benchmark-specific verifiers and predefined scoring rules.
Attributable agent failures are counted as unsuccessful outcomes, while invalid trials and confirmed answer leakage are excluded.
Our primary comparison uses 691 task--model pairs for which all six harnesses have valid scored outcomes.
All reported success rates are computed over this common set.

\subsection{Overall Harness Performance}
\label{sec:exp-overall}

\begin{figure*}[h]
\centering
\includegraphics[width=0.88\textwidth]{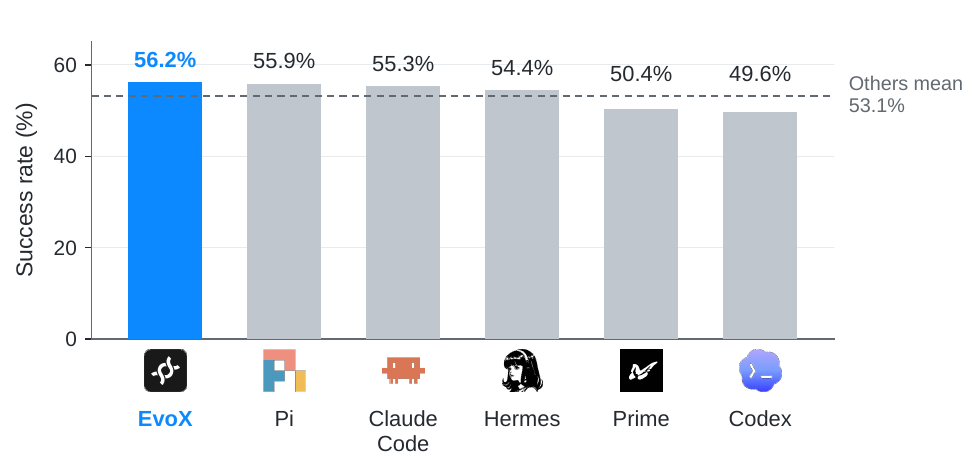}
\caption{
\textbf{Overall performance of six agent harnesses on 691 common task--model pairs across five benchmarks.}
Bars indicate task success rates, with \texttt{EvoX} highlighted in blue.
The dashed line represents the mean success rate of the other five harnesses (53.1\%).
}
\label{fig:exp-overall}
\end{figure*}

\nbf{Competitive Overall Performance.}
As shown in~\cref{fig:exp-overall}, \texttt{EvoX} achieves the highest success rate of \textbf{56.2\%}, followed by Pi (55.9\%), Claude Code (55.3\%), and Hermes (54.4\%).
Prime Agent and Codex achieve 50.4\% and 49.6\%, respectively.
Compared with the mean performance of the other five harnesses (53.1\%), \texttt{EvoX} maintains an advantage of approximately \textbf{3 percentage points}.
Notably, this result is achieved without enabling memory, evolution, or extensions, demonstrating the competitiveness of its baseline operational design.

\nbf{The Role of Harness-Level Control.}
Under matched tasks and foundation models, the observed performance differences highlight the role of harness-level execution in shaping agent outcomes.
While the four leading harnesses achieve relatively close results, \texttt{EvoX} maintains a larger margin over Prime Agent and Codex.
These findings support the importance of the \emph{Operational Policy} as a complementary component to the model-driven \emph{Semantic Policy}, rather than attributing task success solely to foundation-model capabilities.

\subsection{Harness--Model Interactions}
\label{sec:exp-models}

\begin{figure*}[h]
\centering
\includegraphics[width=\textwidth]{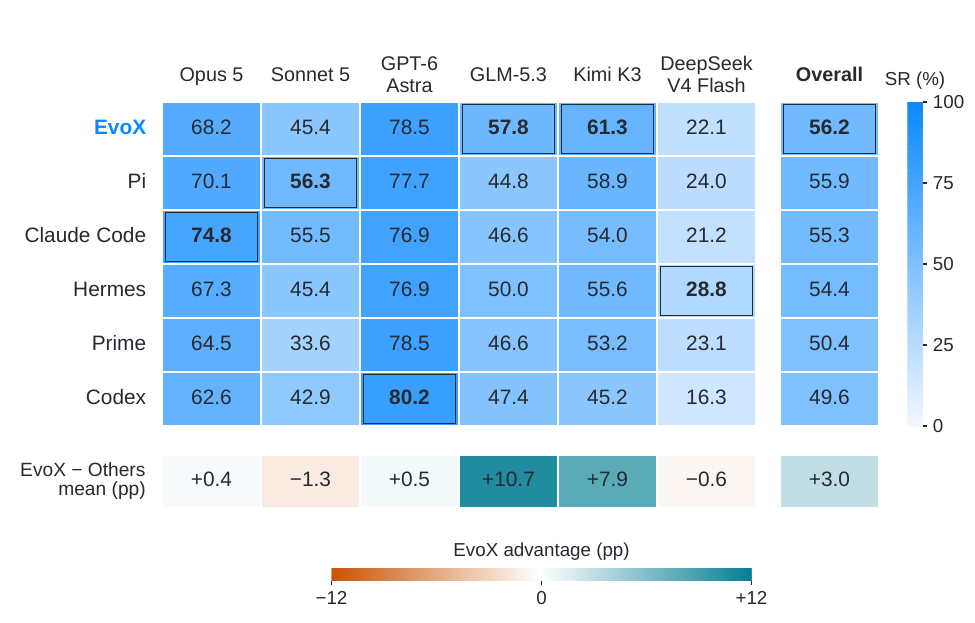}
\caption{
\textbf{Harness--model performance across six foundation models.}
Each cell reports the task success rate (\%) of a harness--model combination, with outlined cells indicating the highest performance for each model.
The overall column summarizes performance across 691 common task--model pairs.
The bottom row reports the performance difference between \texttt{EvoX} and the mean of the other five harnesses, in percentage points.
}
\label{fig:exp-models}
\end{figure*}

As shown in~\cref{fig:exp-models}, harness effectiveness varies substantially across foundation models.

\nbf{Model--Harness Affinity.}
Claude Code achieves the highest success rate on Opus 5 (74.8\%), while Codex leads on GPT-6 Astra (80.2\%), suggesting favorable alignment between these first-party harnesses and their associated model families.
These observations indicate that harness effectiveness depends on the underlying model rather than following a consistent ranking across providers.

\nbf{Cross-Provider Adaptability.}
Unlike provider-specific harnesses, \texttt{EvoX} is designed to operate across heterogeneous foundation-model ecosystems.
It achieves the highest success rates on GLM-5.3 (\textbf{57.8\%}) and Kimi K3 (\textbf{61.3\%}), while remaining competitive on GPT-6 Astra (78.5\%) relative to Codex (80.2\%).
This demonstrates that a provider-independent operational design can effectively support models from different families without relying on a proprietary foundation model.
Nevertheless, \texttt{EvoX} does not consistently lead across all models, highlighting the importance of model-specific compatibility.

\nbf{Model-Conditional Advantages.}
Relative to the mean of the other five harnesses, \texttt{EvoX} improves success rates by \textbf{10.7} percentage points on GLM-5.3 and \textbf{7.9} points on Kimi K3, compared with an overall advantage of approximately \textbf{3.0} points.
Its relative differences are considerably smaller on Opus 5 (+0.4) and GPT-6 Astra (+0.5), and slightly negative on Sonnet 5 (-1.3) and DeepSeek V4 Flash (-0.6).
These variations suggest that the effectiveness of the \emph{Operational Policy} depends in part on the foundation model underlying the \emph{Semantic Policy} it serves.

\subsection{Generalization Across Task Families}
\label{sec:exp-tasks}

\begin{table*}[h]
\centering
\small
\setlength{\tabcolsep}{5pt}
\renewcommand{\arraystretch}{1.08}
\caption{
\textbf{Performance across five benchmark families.}
Each entry reports task success rate (\%) on common task--model pairs, with $n$ denoting the number of pairs per benchmark.
$\Delta$ indicates the difference between \texttt{EvoX} and the mean of the other five harnesses, in percentage points.
Best results are highlighted in bold.
}
\label{tab:exp-tasks}
\begin{tabular}{@{}lcccccccc@{}}
\toprule
Benchmark & \texttt{EvoX} & Pi & Claude Code & Hermes & Prime & Codex & $\Delta$ (pp) \\
\midrule
Aider Polyglot & \textbf{64.7} & 61.7 & 62.3 & 63.5 & 61.1 & 55.1 & +4.0 \\
GAIA & \textbf{86.7} & 85.7 & 82.9 & 85.7 & 85.7 & 78.1 & +3.0 \\
Harbor Index & 27.3 & 25.9 & \textbf{31.5} & 24.5 & 23.1 & 27.3 & +0.8 \\
SWE-Live & 48.4 & \textbf{49.2} & \textbf{49.2} & 43.7 & 35.7 & 37.3 & +5.4 \\
Terminal-Bench 2.1 & 57.3 & \textbf{60.7} & 55.3 & 59.3 & 50.0 & 54.0 & +1.5 \\
\bottomrule
\end{tabular}
\end{table*}

\nbf{Cross-Task Competitiveness.}
To examine whether the competitiveness of \texttt{EvoX} extends across different workloads, \cref{tab:exp-tasks} compares six harnesses on five benchmark families.
\texttt{EvoX} achieves the highest success rates on Aider Polyglot (64.7\%) and GAIA (86.7\%), covering multilingual programming and web-assisted reasoning.
Moreover, \texttt{EvoX} exceeds the average performance of the other five harnesses across all evaluated benchmarks, with gains ranging from 0.8 to 5.4 percentage points.
These results indicate competitive performance across diverse task requirements rather than reliance on a single benchmark.

\nbf{Task-Dependent Performance.}
The relative advantage of \texttt{EvoX} is most pronounced on SWE-Live, where it exceeds the competing harness average by 5.4 percentage points.
However, the strongest individual harness varies across benchmarks: Claude Code leads on Harbor Index, while Pi and Claude Code slightly outperform \texttt{EvoX} on SWE-Live, and Pi leads on Terminal-Bench 2.1.
These differences highlight the absence of a universally dominant harness across task families.
Compared with the pronounced model-dependent variations observed in~\cref{sec:exp-models}, the benchmark-level results suggest that \texttt{EvoX} maintains relatively consistent competitiveness across the evaluated task types.

\subsection{Execution Outcomes and Behavioral Evidence}
\label{sec:exp-behavior}

\begin{table*}[h]
\centering
\small
\setlength{\tabcolsep}{3.5pt}
\renewcommand{\arraystretch}{1.1}
\caption{
\textbf{Execution outcome profiles across 691 common task--model pairs.}
Normal completion excludes nonzero exits, timeouts, and cost-cap interruptions.
Exit denotes a nonzero process exit; TO and Cap denote timeout and cost-cap termination, respectively, with P/F indicating passing or failing verifier outcomes.
A passing verifier result after a timeout is scored as unsuccessful under the evaluation protocol.
The final columns report the normal-completion rate and verifier success rate conditional on normal completion.
}
\label{tab:exp-execution}
\begin{tabular}{@{}lrrrrrrrrrr@{}}
\toprule
& \multicolumn{2}{c}{Normal} & \multicolumn{6}{c}{Other outcomes} & \multicolumn{2}{c}{Rate (\%)} \\
\cmidrule(lr){2-3}
\cmidrule(lr){4-9}
\cmidrule(l){10-11}
Harness & Pass & Fail & Exit & TO-F & TO-P & Cap-P & Cap-F & Other & Normal & Pass$|$Normal \\
\midrule
\texttt{EvoX} & 373 & 134 & 117 & 49 & 4 & 12 & 2 & 0 & 73.4 & \textbf{73.6} \\
Pi & 365 & 187 & 5 & 107 & 6 & 18 & 2 & 1 & \textbf{79.9} & 66.1 \\
Claude Code & 344 & 161 & 111 & 30 & 5 & 37 & 3 & 0 & 73.1 & 68.1 \\
Hermes & 361 & 150 & 159 & 2 & 0 & 14 & 5 & 0 & 74.0 & 70.6 \\
Prime Agent & 330 & 152 & 119 & 68 & 4 & 15 & 3 & 0 & 69.8 & 68.5 \\
Codex & 329 & 154 & 144 & 47 & 3 & 12 & 2 & 0 & 69.9 & 68.1 \\
\bottomrule
\end{tabular}
\end{table*}

\nbf{Execution Outcome Profiles.}
Beyond aggregate success rates, we examine how harnesses terminate their executions and whether completed runs satisfy task objectives.
As shown in~\cref{tab:exp-execution}, \texttt{EvoX} completes 73.4\% of trials normally, closely matching the 73.3\% average of the other five harnesses.
However, among normally completed trials, \texttt{EvoX} achieves a verifier success rate of \textbf{73.6\%}, compared with 68.3\% for the competing harnesses on average.
This suggests that its competitive performance is not simply attributable to a higher frequency of normal termination.
Notably, timeout and exit-code distributions also differ across harnesses, reflecting their distinct execution and termination semantics.

\nbf{Evidence-Grounded Execution.}
To further characterize execution behavior, we examine 22 reviewed \texttt{EvoX} success trajectories alongside comparator runs under matched task--model conditions.
Among 15 selected cases where \texttt{EvoX} succeeds while at least four competing harnesses fail, eight exhibit more extensive execution-grounded verification, including end-to-end testing, differential checks, and failure reproduction.
For example, on \texttt{cancel-async-tasks} with GLM-5.3, \texttt{EvoX} reproduces repeated cancellation through a subprocess receiving real SIGINT signals.
On \texttt{capstone-2743}, it uses GDB to locate an out-of-bounds error and validates the correction through differential checks over approximately 230,000 encodings.
Other trajectories show continued local analysis following unsuccessful attempts to retrieve external information, illustrating how execution can proceed despite intermediate failures.

\nbf{Operational Implications.}
These observations are consistent with the intended roles of \emph{Execution Continuity} and \emph{Outcome Grounding} in the \emph{Operational Policy}.
Beyond producing a final response, \texttt{EvoX} exhibits behaviors that incorporate execution feedback and verification into task completion.

\subsection{Evaluation Integrity and Sensitivity}
\label{sec:exp-reliability}

\begin{figure*}[h]
\centering
\includegraphics[width=\textwidth]{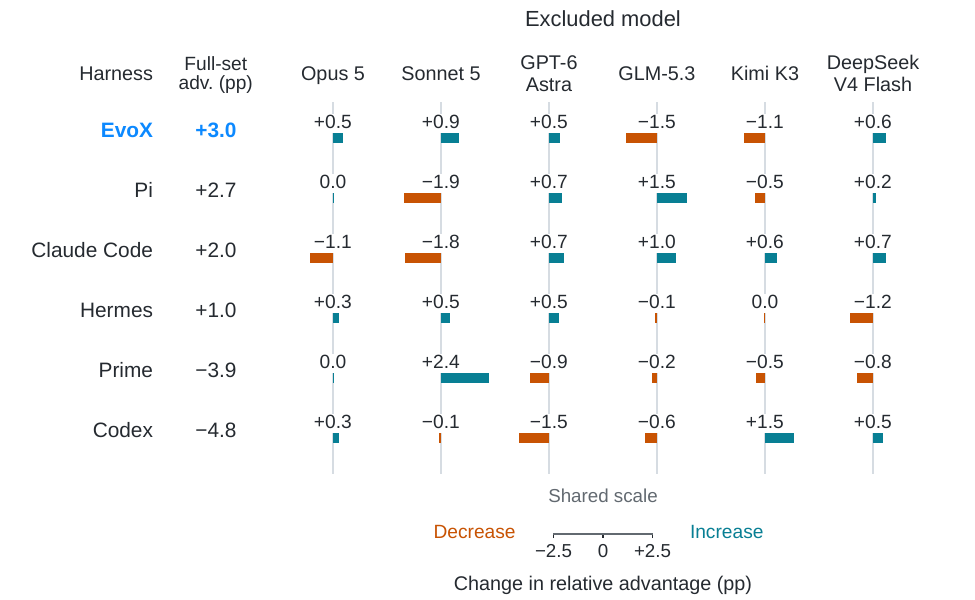}
\caption{
\textbf{Model-composition sensitivity across six agent harnesses.}
The first column reports each harness's performance advantage over the mean of the other five harnesses on the full evaluation set.
Each remaining entry shows the change in this relative advantage after excluding the indicated foundation model, measured in percentage points.
Negative values indicate reduced relative competitiveness, while positive values indicate improvement.
}
\label{fig:exp-reliability}
\end{figure*}

\nbf{Cross-Harness Model Sensitivity.}
We examine how the relative competitiveness of each harness changes when individual foundation models are excluded from evaluation.
As shown in~\cref{fig:exp-reliability}, different harnesses exhibit distinct sensitivity patterns.
Excluding Sonnet 5 reduces the relative advantage of Pi and Claude Code by 1.9 and 1.8 percentage points, respectively, while excluding GPT-6 Astra reduces that of Codex by 1.5 points.
In contrast, Prime Agent gains 2.4 points when Sonnet 5 is excluded.
These variations further illustrate that harness effectiveness depends in part on the foundation models it serves.

\nbf{EvoX Across Model Subsets.}
With all six models, \texttt{EvoX} exceeds the average of the other five harnesses by approximately 3.0 percentage points.
Its advantage remains positive under every single-model exclusion, indicating competitive performance across the examined model subsets.
However, excluding GLM-5.3 reduces its advantage by 1.5 points, while excluding Kimi K3 reduces it by 1.1 points.
These results reinforce the importance of model--harness compatibility while highlighting the contribution of GLM-5.3 and Kimi K3 to \texttt{EvoX}'s aggregate performance.

\nbf{Evaluation Integrity and Scope.}
To ensure evaluation integrity, \texttt{Contrast} examines task environments and execution records for potential answer leakage, excluding cases confirmed through review.
Our evaluation covers 691 common task--model pairs across five benchmarks, six foundation models, and six agent harnesses under matched evaluation conditions.
For \texttt{EvoX}, the reported results characterize its baseline operational behavior, with memory, evolution, and extensions disabled.

\section{Conclusion}
\label{sec:conclusion}

In this work, we present \texttt{EvoX} and the perspective of \emph{The Second Policy}, viewing the agent harness as an \emph{Operational Policy} that complements the foundation model's \emph{Semantic Policy}.
\texttt{EvoX} realizes this perspective through action realization, execution continuity, and outcome grounding, while \texttt{Contrast} enables controlled and auditable comparisons across heterogeneous harnesses.
Across five benchmarks, six foundation models, and six harnesses, \texttt{EvoX} achieves the highest numerical success rate of 56.2\%, exceeding the competing harness average by approximately 3.0 percentage points.
Its competitive cross-provider performance, particularly on GLM-5.3 and Kimi K3, highlights the importance of harness--model compatibility.
These results are achieved without memory, evolution, or extensions, establishing a competitive foundation for further operational adaptation.
Looking ahead, we envision the harness as a promising substrate for agent self-evolution, where verified execution experience can be accumulated, refined, and reused to improve future behavior.
Through experience sharing and coordination across agents, this direction could further extend toward collective evolution and self-organizing agent swarms.
Together, these findings and possibilities highlight the broader potential of harness design for building reliable, adaptive, and collaboratively evolving agent systems.

\bibliographystyle{unsrtnat}
\bibliography{references}

\end{document}